\documentclass[conference]{IEEEtran}
\IEEEoverridecommandlockouts
\usepackage[left=15.9mm,right=15.9mm,top=19.1mm,bottom=26.5mm]{geometry} 
\usepackage{cite}
\usepackage{amsmath,amssymb,amsfonts}
\usepackage{graphicx}
\usepackage{booktabs}
\usepackage{textcomp}
\usepackage{xcolor}
\usepackage{array}
\usepackage{multirow}
\usepackage{float}
\usepackage{caption}
\usepackage{subcaption}
\usepackage{footnote}
\usepackage{hyperref}
\usepackage{hyperref} 
\hypersetup{
    linkbordercolor={0 0 1}
}
\usepackage{tikz}
\usetikzlibrary{shapes.geometric, arrows}

\usepackage{algorithm}
\usepackage{algpseudocode}

\newcolumntype{P}[1]{>{\centering\arraybackslash}p{#1}}

\newcommand{\rom}[1]{\uppercase\expandafter{\romannumeral #1\relax}}

\newcommand{\figref}[1]{Fig.~\ref{#1}}

\def\BibTeX{{\rm B\kern-.05em{\sc i\kern-.025em b}\kern-.08em
    T\kern-.1667em\lower.7ex\hbox{E}\kern-.125emX}}

\begin{document}
\title{Real-Time AI-Driven People Tracking and Counting Using Overhead Cameras}
\author{Ishrath Ahamed$^1$\IEEEauthorrefmark{1}, Chamith Dilshan Ranathunga$^1$\IEEEauthorrefmark{1}, Dinuka Sandun Udayantha$^1$\IEEEauthorrefmark{2},\\
Benny Kai Kiat Ng$^2$, and Chau Yuen$^3$ \textit{Fellow, IEEE}\\[5pt]
\thanks{\IEEEauthorrefmark{1}Co-first authors. These authors contributed equally to this work.}
\thanks{\IEEEauthorrefmark{2}Second author.}
$^1$Dept. of Electronic and Telecommunication Engineering, University of Moratuwa, Sri Lanka\\
$^2$Engineering Product Development Pillar, Singapore University of Technology and Design, Singapore\\
$^3$School of Electronic and Electrical Engineering, Nanyang Technological University, Singapore
\vspace{-15pt}
\thanks{\copyright 20XX IEEE.  Personal use of this material is permitted.  Permission from IEEE must be obtained for all other uses, in any current or future media, including reprinting/republishing this material for advertising or promotional purposes, creating new collective works, for resale or redistribution to servers or lists, or reuse of any copyrighted component of this work in other works.}
}

\maketitle

\begin{abstract}
Accurate people counting in smart buildings and intelligent transportation systems is crucial for energy management, safety protocols, and resource allocation. This is especially critical during emergencies, where precise occupant counts are vital for safe evacuation. Existing methods struggle with large crowds, often losing accuracy with even a few additional people. To address this limitation, this study proposes a novel approach combining a new object tracking algorithm, a novel counting algorithm, and a fine-tuned object detection model. This method achieves 97\% accuracy in real-time people counting with a frame rate of 20-27 FPS on a low-power edge computer.

\end{abstract}


\section{INTRODUCTION}
\label{sec:intro}

Keeping track of the number of people who have entered, exited, and are remaining inside a building or public transport is essential for crowd, facility, and safety management. Even though recent advancements in technology have paved the path for smart buildings and intelligent public transportation systems, it remains challenging to develop economical, efficient, and accurate systems to count people in real-time. Several works have been conducted since the early 2000s for real-time people counting using signal processing and deep learning techniques. Regardless of the technology they use, the reliability, accuracy of these, and cost to implement them are questionable. Either the highly accurate systems are extremely expensive or low-cost systems are not accurate and reliable.

The use of time-of-flight concept sensors such as laser beam, thermal, and ultrasound sensors tends to fail when two or more people need to be counted. Jae Hoon \textit{et al.} \cite{jae_hoon_lee_kim_bong_keun_kim_k._ohba_kawata_akihisa_ohya_sasaki_2007} attempted using laser range finders. The system requires two sensors to function properly, and these sensors need specific placement depending on the environment. This customization makes it difficult to implement in various settings. In order to address these issues Jeong Woo \textit{et al.} \cite{choi_quan_cho_2018} proposed a new method using two ultra-wide band radar sensors. Even in this approach, the placement of two sensors is crucial, placing them very close leads to undercounting, and placing them far apart leads to miscalculations such as double counting.
Yanni \textit{et al.} \cite{yang_cao_liu_liu_2019} and Tiang \textit{et al.} \cite{tian_chen_xu_chen_2021} have done studies employing WiFi and analyzing the phase, but they are not ideal as WiFi easily get distorted by other objects leading the system for inaccurate results.

In this deep learning and image processing era, several works have been carried out to count people. In \cite{cao_sun_odoom_luan_song_2016}, they introduced an image processing-based method employing background subtraction and blob detection to track objects, but it suffered from sensitivity to illumination changes and false positives caused by non-human objects in the frame. Li Guangqin \textit{et al.}~\cite{li_ren_lyu_zhang_2016} proposed a depth camera-based solution to mitigate illumination issues but still faced challenges with false positives, while \cite{sun_akhtar_song_zhang_li_mian_2019}, inspired by \cite{li_ren_lyu_zhang_2016}, used depth cameras for background subtraction and 3D reconstruction for person detection, yet struggled with detecting individuals outside predefined human model dimensions. In response to challenges with traditional image processing methods, researchers have explored deep learning approaches for person detection and tracking. Guojin \textit{et al.} \cite{liu_yin_jia_xie_2017} utilizes a convolutional neural network (CNN) for head detection and a spatio-temporal context tracking algorithm but faces issues with computational efficiency and sensitivity to head rotations, while \cite{gamanayake_jayasinghe_ng_yuen_2020} proposes ``cluster pruning" to enhance real-time performance, achieving improved frames per second (FPS) values for people counting tasks but still falling short for monitoring door crossings due to low FPS rates.

Therefore, a low-cost yet reliable system with a high FPS rate is needed for real-time people counting on edge devices. To this end, we propose a new efficient tracking algorithm, a counting algorithm, and a fine-tuned model for object detection in any complex environment. With these proposed methods we obtain higher FPS rates even when there are more than two people in the frame in both good and low lighting conditions. This achieves an overall accuracy of 97\% in real-time video testing which is a 2\% improvement compared to the best state-of-the-art and high frame rate: 20-27 FPS on average.
\section{METHODOLOGY}
\label{sec:method}
\begin{figure*}[ht]
  \begin{subfigure}{0.33\textwidth}
    \centering
    \includegraphics[width=\linewidth]{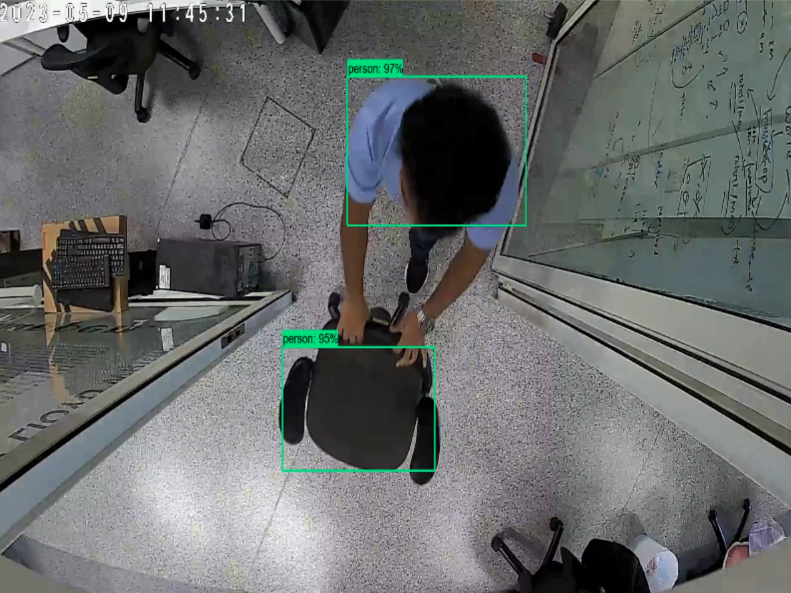}
    \caption{Before specifically training our model to avoid potential distractions}
  \end{subfigure}
  \begin{subfigure}{0.33\textwidth}
    \centering
    \includegraphics[width=\linewidth]{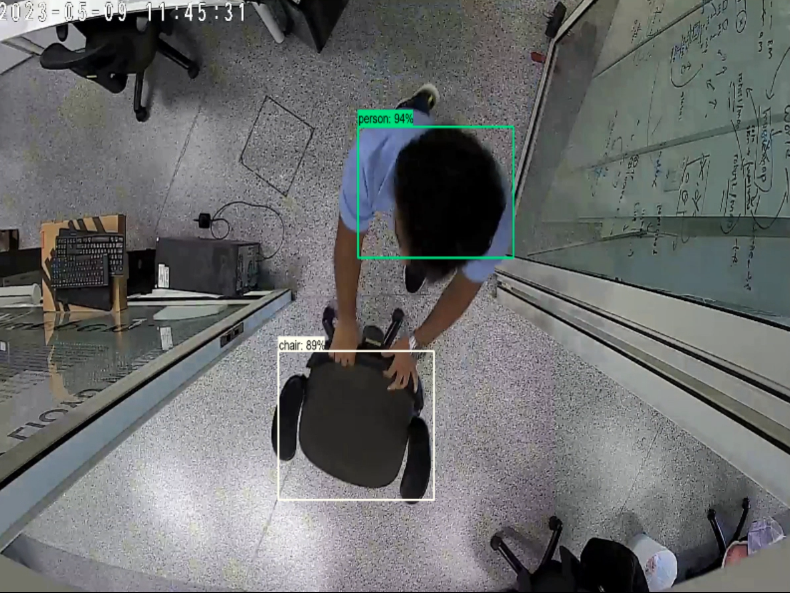}
    \caption{After training our model to avoid potential distractions}
  \end{subfigure}
  \begin{subfigure}{0.33\textwidth}
      \centering
      \includegraphics[width=\linewidth]{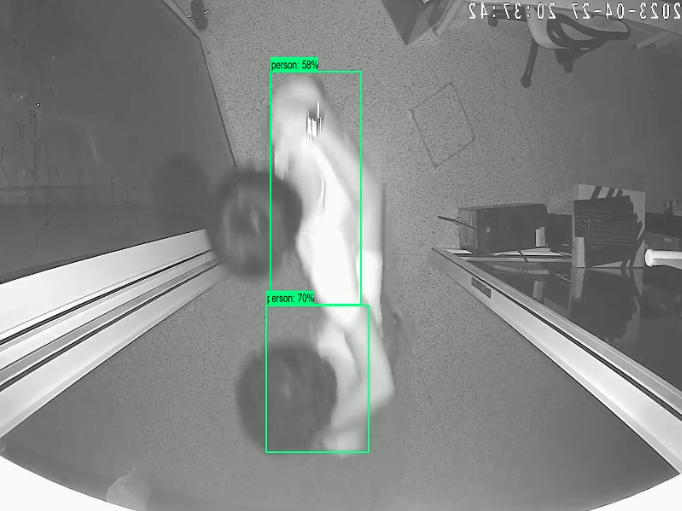}
      \caption{After training our model for low-light conditions}
  \label{fig:People detection in low/no light conditions}
  \end{subfigure}
  \caption{Before specifically training our model to identify potential disturbances, there were few mismatches, identifying distractions as potential head objects. For example in \textbf{(a)} the chair is identified as a person with 95\%. But after training \textbf{(b)} the chair is identified as a chair with 89\% confidence, while the person is identified with 97\% \& 94\% confidence in both cases respectively.\\ \textbf{(c)} people detection in low/no light conditions. The model correctly detects the two people in the IR camera frame with 70\% (lower) and 58\% (top) confidence.}
  \vspace{-10pt}
  \label{fig:disturbances} 
\end{figure*}

This section is divided into four subsections. In the subsection~\ref{sec:detection}, we introduce the the proposed object detection model. Subsections \ref{sec:extraction} and~\ref{sec:tracking} are dedicated to discuss the newly introduced feature extraction method and tracking algorithm respectively. Finally in subsection~\ref{sec:counting} we introduce the method employed in this proposed algorithm to count people.

\subsection{Head Detection}\label{sec:detection}

Our video input originates from an overhead camera positioned above a doorway. We initially consider background subtraction and blob detection for head identification, as explored by prior researchers. However, this method suffers from accuracy limitations, further exacerbated by complexities introduced by the opening and closing of the door. Additionally, these methods present significant challenges in distinguishing heads from potential distractions such as bags, chairs, and trolleys.

Consequently, we opt for object detection using deep learning techniques to identify human heads. We identify that the SSD Mobilenet model offers a compelling combination of accuracy, speed, and adaptability for our requirements.\cite{DBLP:journals/corr/LiuAESR15}

Our head detection with SSD Mobilenet employs two distinct, self-trained models:
\begin{enumerate}
    \item Performs object detection under daylight conditions.
    \item Performs object detection in low-light/night light conditions.
\end{enumerate}


The camera we use (\href{https://www.tapo.com/en/product/smart-camera/tapo-c100/}{(Tapo C100 WiFi Camera)} is equipped with infrared (IR) mode to automatically discern between daytime and nighttime conditions. During nighttime, the camera switches to IR mode, resulting in grayscale images where the pixel values across all three channels (R, G, and B) remain constant as in Eq.~\ref{eq:rgb_equal}. Therefore, we employ this characteristic of IR mode to determine nighttime conditions. When the camera operates in IR mode, indicating nighttime, the second model is utilized for object detection, ensuring optimal performance in low-light environments. One such instance is visualized in \figref{fig:People detection in low/no light conditions}
\begin{align}
    \forall \, \text{pixels} \ (x, y); \quad R(x, y) &= G(x, y) = B(x, y)
    \label{eq:rgb_equal}
\end{align}


Our efforts focus on mitigating the influence of potential distractions, including chairs, trolleys, and bags. To achieve this, we fine-tune the SSD Mobilenet model to not only identify human heads captured by the overhead camera but also to distinguish them from these three aforementioned distractions. In \figref{fig:disturbances}, such an instance with a chair is visualized.

\subsection{Feature Extraction}\label{sec:extraction}

To enable head tracking across video frames, we extract features from the detected heads in each frame. These features provide a robust representation of the head, facilitating its identification in subsequent frames. We employ the MobileNetV2 \cite{mobilenet} model for feature extraction due to its advantageous combination of lightweight architecture, accuracy, and mobile device compatibility. Notably, MobileNetV2 is pre-trained on the vast ImageNet dataset, offering a strong foundation for image matching tasks in our application.

We identify the feature extraction block as the system's computational bottleneck. To expedite the feature extraction process, we implement a two-pronged optimization strategies:

 \begin{itemize}
     \item \textbf{Model Compression:} We compressed the MobileNetV2 model into a ``.tflite" format. This significantly reduces the model size, leading to faster loading and execution times.

     \item \textbf{Input Downsampling:} We downsampled the input image fed to the feature extraction model to a resolution of (120, 120, 3). This technique effectively reduces the computational burden associated with feature extraction. 
 \end{itemize}

 Consequently, the input provided to the MobileNetV2 model is a cropped head image of size (120, 120, 3), and the model outputs an embedding vector of size (1, 1024).
\subsection{Tracking}\label{sec:tracking}
Following head detection and feature extraction in each video frame, we implement a custom-designed tracking algorithm to facilitate person counting by monitoring head movement across subsequent frames. This algorithm functions by assigning a unique identifier (ID) to each newly detected head object. In subsequent frames, the algorithm compares the extracted features of newly detected head objects with those associated with existing head objects. If the features exhibit significant similarity, the newly detected head object is assigned the same ID as the corresponding existing head object. Conversely, if the features demonstrate a low degree of similarity, a new ID is assigned to the newly detected head object.

The object tracking algorithm is extensively outlined in Algorithm \ref{alg:object tracking}. $M$ is the distance matrix that contains the feature distances between newly detected objects in the current frame and already registered objects in the program. Matrix element $M_{i,j}$ denotes the feature distance between the $i^{th}$ registered object and the $j^{th}$ newly detected object. Here, $i \leq m$ and $j \leq n$, where $m$ and $n$ are the numbers of already registered objects and newly detected objects, respectively. Now, the object tracking problem is reduced to an assignment problem with the objective of minimizing the feature distance cost. The number of maximum possible assignments and assignments done is denoted by $A$ and $a$, respectively, where $A = \min(m, n)$. The threshold for feature distance and spatial distance is denoted by $T$ and $D$, respectively. Let us also define the spatial distance matrix $N$. We also allow some degree of detection misses by setting a property called \texttt{eCount} for the objects. Let us denote the threshold for the number of consecutive detection misses for a particular object by $E$.
\begin{algorithm}
\vspace{-2pt}
\caption{Object tracking}\label{alg:object tracking}
\begin{algorithmic}[1]
\State $a \gets 0$
\While {$(a < A)$ \textbf{and} ($(\min(M) < T)$)}
    \State $(i, j) \gets \arg\min_{i,j} M_{i,j}$
    \If {($i$ \textbf{or} $j$ already assigned) \textbf{or} ($N_{i,j} > D$)}
        \State $M_{i,j} \gets T$
        \State go to next iteration
    \EndIf
    \State $i \gets j$
    \State $i.\texttt{eCount} \gets 0$
    \State $M_{i,j} \gets T$
    \State $a \gets a + 1$ 
\EndWhile
\State $\forall j\text{, }  j\texttt{.ID} \gets \texttt{newID} \text{; } j \in \{\text{unassigned new objects}\}$
\State $\forall i \text{, } i.\texttt{eCount} \mathrel{+}= 1 \text{; } i \in \{\text{unassigned old objects}\}$
\For{each $object$ in the program}
    \If {$(object.\texttt{eCount} > E)$}
        \State remove $object$ from program
    \EndIf
\EndFor
\end{algorithmic}
\vspace{-2pt}
\end{algorithm}

The complete pipeline of our methodology is depicted in Figure \ref{fig:object_tracking_pipeline}. This hand-crafted tracking approach offers a balance of simplicity and efficiency in real-time applications. Moreover, its reliance on feature-based tracking, rather than solely on bounding boxes, enhances robustness against challenges such as occlusion that are commonly encountered in real-world video data.
\definecolor{white}{RGB}{255, 255, 255}
\definecolor{black}{RGB}{0, 0, 0}

\tikzstyle{box_style} = [rectangle, rounded corners, minimum width=3cm, minimum height=0.8cm, text centered, font=\normalsize, color=black, draw=black, line width=1, fill=white]
\tikzstyle{thick_arrow} = [thick, draw=black, line width=1, ->, >=stealth]

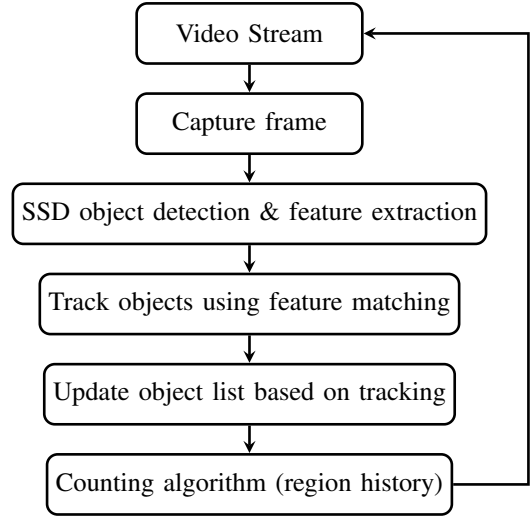
\begin{figure}[!t]
  \centering
  \begin{tikzpicture}[node distance=1cm]
    \node (video_stream) [box_style] {Video Stream};
    \node (capture_frame) [box_style, below of=video_stream, yshift=-0.2cm] {Capture frame};
    \node (object_detection) [box_style, below of=capture_frame, yshift=-0.2cm] {SSD object detection \& feature extraction};
    \node (tracking) [box_style, below of=object_detection, yshift=-0.2cm] {Track objects using feature matching};
    \node (delete_untracked) [box_style, below of=tracking, yshift=-0.2cm] {Update object list based on tracking};
    \node (counting_algorithm) [box_style, below of=delete_untracked, yshift=-0.2cm] {Counting algorithm (region history)};

    \draw [thick_arrow] (video_stream) -- (capture_frame);
    \draw [thick_arrow] (capture_frame) -- (object_detection);
    \draw [thick_arrow] (object_detection) -- (tracking);
    \draw [thick_arrow] (tracking) -- (delete_untracked);
    \draw [thick_arrow] (delete_untracked) -- (counting_algorithm);
    \draw [thick_arrow] (counting_algorithm.east) -- ++(1,0) |- (video_stream.east);
  \end{tikzpicture}
  \caption{Object Tracking Pipeline}
  \label{fig:object_tracking_pipeline}
\end{figure}

\subsection{Counting Algorithm}\label{sec:counting}
The field of view (FOV) obtained by the overhead camera is divided into 3 regions, separated by 2 horizontal lines:
\begin{itemize}
    \item A: Totally outside region
    \item B: Critical region
    \item C: Totally inside region
\end{itemize}
To facilitate person counting, we incorporate a region history attribute for each ``person-object." Dividing the field of view into three distinct regions surpasses the limitations of two-region systems that are susceptible to disturbing oscillations when people stay near the central boundary line. This region history attribute maintains a record of the object's traversal path. When a person-object moves from Region A to Region C, it is registered as entering the designated space (room). Conversely, movement from Region C to Region A signifies exiting the space.

\section{RESULTS AND DISCUSSION}
\label{sec:results}
\begin{figure*}[!ht]
\hspace{27pt}
  \begin{subfigure}{0.41\textwidth}
    \centering
    \includegraphics[width=\linewidth]{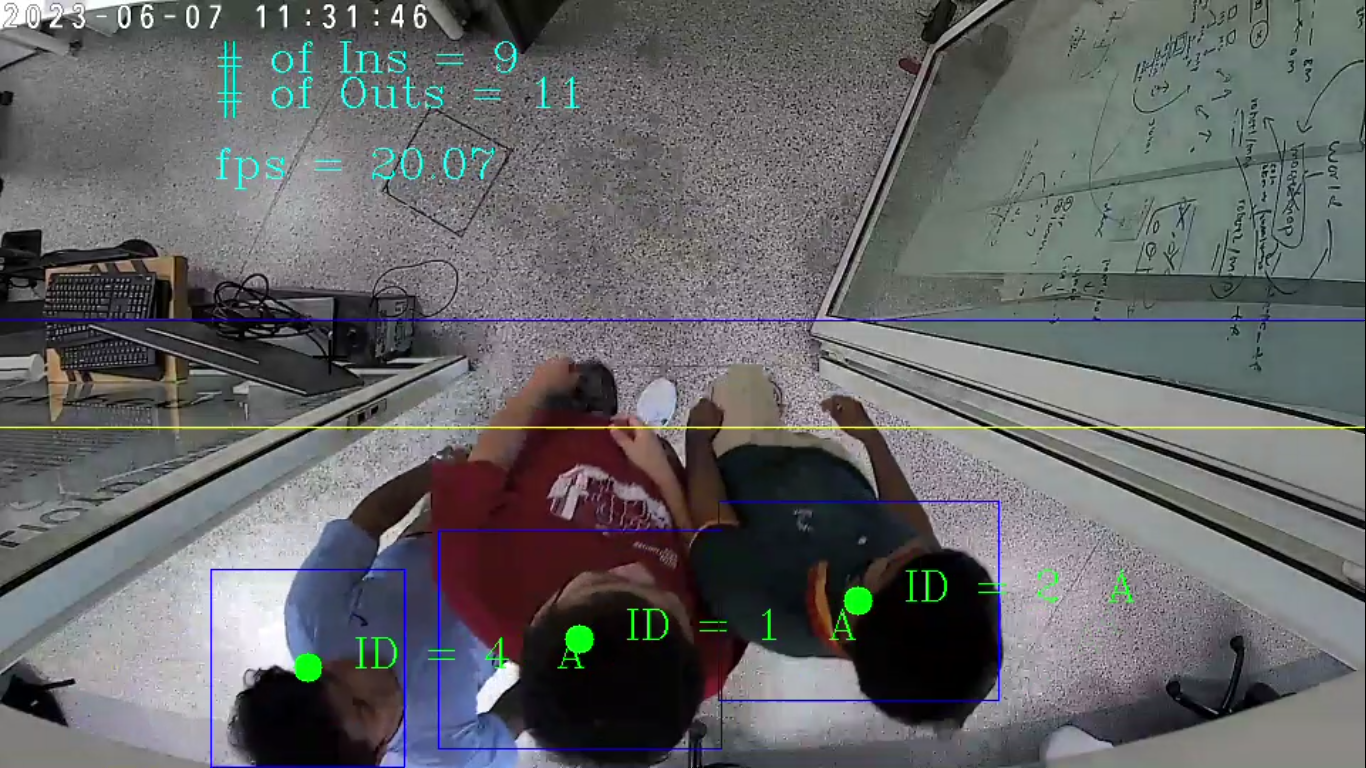}
    \caption{}
  \end{subfigure}
  \hfill
  \begin{subfigure}{0.41\textwidth}
    \centering
    \includegraphics[width=\linewidth]{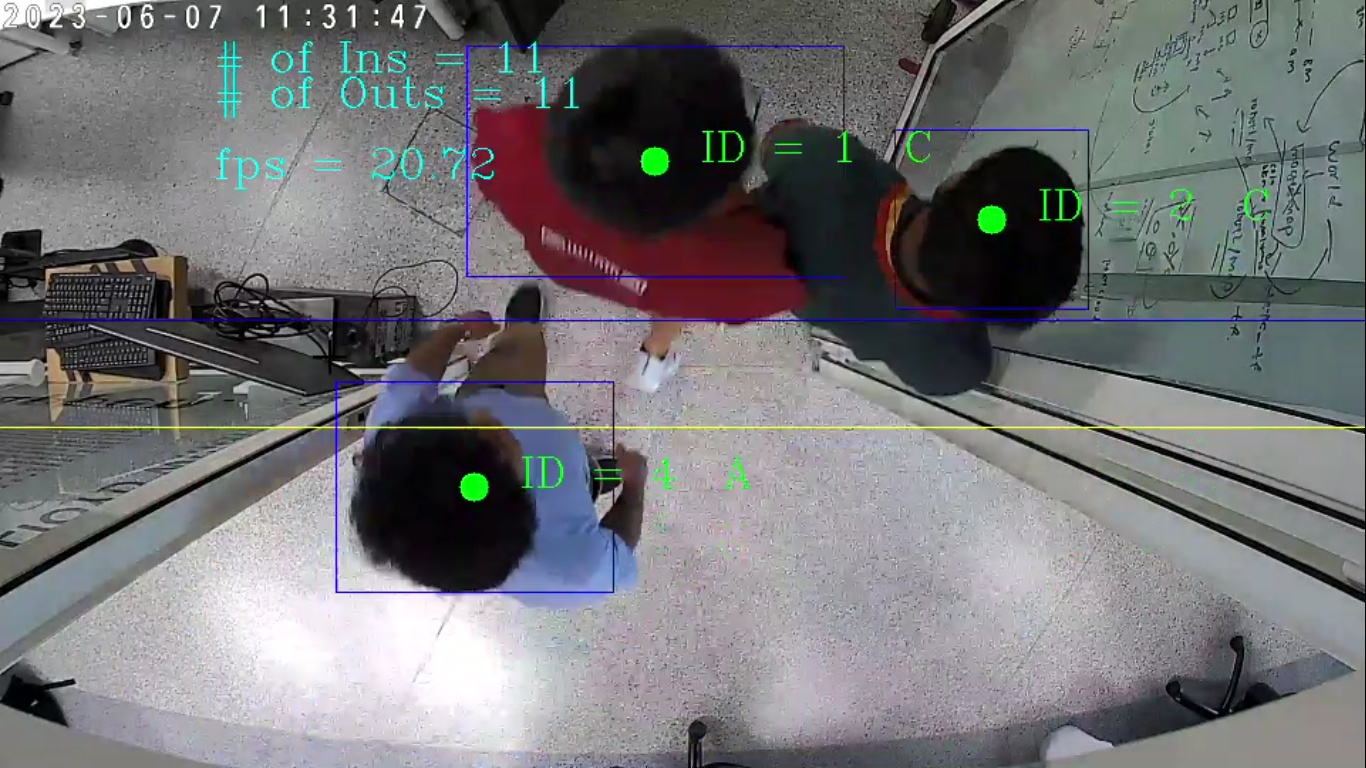}
    \caption{}
  \end{subfigure}
  \hspace{27pt}
  \caption{Real time deployment result can be found here \cite{real_time_results}: (a) before entering the room. ``\# of Ins" is 9. (b) After two people, the man wearing a red T-shirt and the man wearing a green T-shirt, entering the room, ``\# of Ins" has been increased by 2 indicating 2 people entered the room. Also their assigned IDs change from A to C indicating the shift from outside to inside.}
  \label{fig:Real time result}
  \vspace{-10pt}
\end{figure*}

Our model is successfully implemented on an Intel® NUC 12 Pro Mini PC, the NUC12WSHi7 model. No GPU is used in the testing stage. The processor has 12 cores, 16 threads, and a max turbo frequency of up to 4.7 GHz in burst mode. The frame rate remains consistently above 23 frames per second (FPS) under standard conditions. Even with an increased number of individuals present in the frame, there is only a slight drop in FPS, which remains at an acceptable level for real-time applications. Table \ref{table:average_fps} presents the average FPS observed under different scenarios, and Fig.~\ref{fig:Real time result} shows the results of one real-time instance where three people are present in the camera frame at the same time.

\begin{table}[htbp]
    \centering
    \caption{Average FPS with Varying Number of People}
    \label{table:average_fps}
    \begin{tabular}{|c|c|c|c|c|}
        \hline
        \textbf{Number of People} & 0 & 1 & 2 & 3 \\
        \hline
        \textbf{Average FPS} & 27 & 25 & 22 & 20 \\
        \hline
    \end{tabular}
\end{table}

Live video testing is conducted to evaluate the model's effectiveness in real-world scenarios. During this test, combinations of instances of individuals entering and exiting the frame are captured. The model achieves an overall accuracy of 97\%. This demonstrates its robustness in dynamic environments. Furthermore, the model undergoes rigorous testing in a controlled laboratory setting, continuously monitoring a room for two consecutive days. Table \ref{table:results_long_term} summarizes the detailed results, highlighting the model's stability and reliability over extended periods. Accuracy is calculated using Eq~\ref{eq:accuracy}.
\begin{equation}
    \label{eq:accuracy}
    \text{Accuracy (\%)} = \frac{\text{Total observations - Error}}{\text{Total observations}} \times 100
\end{equation}

\begin{table}[htbp]
    \centering
    \caption{Long-Term Monitoring Results}
    \label{table:results_long_term}
    \begin{tabular}{|l|c|c|c|c|}
        \hline
        \textbf{Day} & \multicolumn{2}{c|}{\textbf{Number of In / Outs}} & \textbf{Error} & \textbf{Accuracy (\%)} \\
        \cline{2-3}
         & \textbf{Actual} & \textbf{Predicted} & & \\
        \hline
        Day 1 & 15 / 14 & 15 / 17 & 3 & 89.66 \\
        Day 2 & 11 / 10 & 11 / 9 & 1 & 95.24 \\
        \hline
        Overall & 26 / 24 & 26 / 26 & 4 & 92.00 \\
        \hline
    \end{tabular}
\end{table}

Previous attempts by other researchers have shown a trade-off between accuracy and FPS. When accuracy increases, the FPS decreases significantly as shown in table \ref{table:comparison}. However, our proposed solution offers higher accuracy without compromising FPS, making it suitable for live deployments.

\begin{table}[h]
\centering
\caption{Comparison of our method with previous methods}
\begin{tabular}{|l|c|c|}
\hline
\textbf{Methods} & \textbf{Accuracy} & \textbf{FPS} \\ \hline
Shijie \textit{et al.} \cite{sun_akhtar_song_zhang_li_mian_2019}    & 86.32\%          & 45.0        \\ \hline
Guojin \textit{et al.} \cite{liu_yin_jia_xie_2017}   & 93.10\%          & 0.6         \\ \hline
Chinthaka \textit{et al.} \cite{gamanayake_jayasinghe_ng_yuen_2020} & 95.00\%          & 5.0         \\ \hline
Ours                     & \textbf{97.00\%} & \textbf{23} \\ \hline
\end{tabular}
\label{table:comparison}
\end{table}


\section{CONCLUSION}
\label{sec:conclusion}


Despite the advancements in deep learning and image processing, accurately counting people from an overhead camera with low computational power has remained a challenge. In response, we have developed a new algorithm that leverages deep learning and image processing techniques to achieve state-of-the-art results in real-time people counting, regardless of lighting conditions. This algorithm can achieve impressive FPS ranging from 20 to 27 due to a highly optimized object tracking algorithm and fine-tuned object detection model. Future work includes improving the FPS rate for more than 3 people by optimizing the feature extraction method.


\bibliographystyle{IEEEtran}
\bibliography{refs}

\end{document}